\pdfoutput=1
\documentclass[11pt]{article}

\usepackage[letterpaper,margin=1in]{geometry}
\usepackage[T1]{fontenc}
\usepackage[utf8]{inputenc}
\usepackage[english]{babel}
\usepackage{newtxtext,newtxmath}
\usepackage{microtype}
\usepackage[round,authoryear]{natbib}
\usepackage[table]{xcolor}
\usepackage{tabularx}
\usepackage{amsmath}
\usepackage{booktabs}
\usepackage{multirow}
\usepackage{graphicx}
\usepackage{array}
\usepackage{siunitx}
\usepackage{float}
\usepackage{placeins}
\usepackage{url}
\usepackage{xurl}
\usepackage[
    colorlinks=true,
    linkcolor=blue!60!black,
    citecolor=blue!60!black,
    urlcolor=blue!60!black,
    bookmarksnumbered,
    hypertexnames=false
]{hyperref}
\usepackage{bookmark}
\usepackage{authblk}

\newenvironment{theappendices}{\appendix}{}
\newcommand{\chapter}{\section}

\title{Synthetic Crystal Data Generation for Diffusion Pretraining in Material Property Prediction}
\author{Aamod Varma}
\author{Shuyi Jia}
\author{Victor Fung}
\affil{Department of Computer Science, Georgia Institute of Technology, Atlanta, GA 30332\protect\\
\texttt{\{aamodvarma, shuyijia, victorfung\}@gatech.edu}}
\date{}

\hypersetup{
  pdftitle={Synthetic Crystal Data Generation for Diffusion Pretraining in Material Property Prediction},
  pdfauthor={Aamod Varma, Shuyi Jia, Victor Fung}
}

\begin{document}

\maketitle
\begin{abstract}
{Diffusion-based pretraining has emerged as a powerful strategy for learning transferable representations in graph neural networks for materials property prediction without requiring quantum-mechanical labels such as energies, forces, and stresses for every pretraining structure. However, the method retains a critical dependency: it requires structures residing near local minima of a potential energy surface, which have traditionally been obtained through costly DFT relaxations of known or physically synthesized compounds. This reliance constrains both the scale and the structural diversity of available pretraining data, leaving large regions of crystal structure space unexplored. In this work, we connect diffusion-based pretraining with synthetic structure generation to reduce this bottleneck. We introduce a synthetic data generation pipeline combining three complementary strategies: perturb-and-relax, generative sampling via MatterGen, and protostructure enumeration. Where structural relaxation is required, we use fast machine-learned interatomic potentials instead of performing a new DFT relaxation for each generated structure. We further compare a coordinate-denoising control with continuous-time variance-preserving (VP) and sub-VP diffusion objectives using a randomly initialized Orb-v2 backbone. The source comparison covers seven MatBench property-prediction tasks, and an eighth task, \texttt{mp\_gap}, is included in the protostructure scaling study. Every synthetic-pretraining condition numerically outperforms random initialization on all seven source-comparison tasks. Scaling the protostructure corpus from 20k to 165k under a fixed 400-epoch schedule numerically matches or exceeds MP-20k pretraining on most tasks; because this protocol increases both data diversity and optimization updates, the result reflects joint data-and-compute scaling.}
\end{abstract}

\section{Introduction}

Predicting material properties is a central challenge in computational materials science, underpinning the discovery and design of catalysts, batteries, electronic materials, and other functional compounds \citep{pilania2013accelerating, ramprasad2017machine, schmidt2019recent, chien2024opportunities}. To address this challenge at lower computational cost than quantum-mechanical methods such as density functional theory (DFT), graph neural networks (GNNs) have emerged as powerful models for atomistic systems, representing structures as graphs and learning from interactions between neighboring atoms \citep{gilmer_neural_2017, xie2018crystal}. Their ability to efficiently capture complex structure–property relationships has made GNNs the backbone of many state-of-the-art machine-learning interatomic potentials (MLIPs) \citep{yang2024mattersim, horton2025accelerated, zhang2025roadmap, wood2026family}. More recently, pretraining GNNs on large atomistic datasets has emerged as a promising strategy for learning transferable representations that further improve downstream property prediction \citep{zaidi_pre-training_2022, shoghi_molecules_2023, jia_derivative-based_2024, feng2026ssformer}.

Pretraining strategies for GNNs for materials property prediction can be broadly divided into two categories: supervised and self-supervised. Supervised approaches typically leverage large datasets labeled with quantum-mechanical quantities such as energies, forces, and stresses \citep{shoghi_molecules_2023, jia_derivative-based_2024}, or with chemically informed descriptors and fingerprints \citep{jia2025pre}, enabling GNNs to learn chemically and geometrically meaningful representations that transfer effectively to downstream tasks \citep{yamada2019predicting, gupta2021cross}. In contrast, self-supervised approaches require no explicit property labels, instead constructing training targets directly from atomic structures \citep{wang2022molecular, song_diffusion-based_2024, zhang2025cl}. For example, \cite{zaidi_pre-training_2022} perturb equilibrium atomic coordinates with Gaussian noise and pretrain a GNN to recover the added perturbations, an objective that can be interpreted as learning an approximate force field from equilibrium structures. Building on this denoising framework, \cite{neumann_orb_2024} formulate pretraining as a multi-scale diffusion process, perturbing ground-state structures at multiple noise levels and training the model to predict the corresponding noise or score before supervised fine-tuning on energies, forces, and stresses.

Although denoising and diffusion-based pretraining can remove the need for explicit energy, force, or stress labels, their data requirements can still limit scalability and structural diversity.  Conventional denoising objectives typically use equilibrium or near-equilibrium structures, for which restoring perturbed coordinates can be connected to learning the local potential energy landscape \citep{wang2023denoise, wu2026siamese}. However, this restriction is not universal. Denoising non-equilibrium structures (DeNS) extends denoising to off-equilibrium configurations by conditioning on the force of the original structures, thus improving equivariant force fields \citep{liao2024generalizingdenoisingnonequilibriumstructures}. However, its reliance on force labels means that this does not eliminate  the need for quantum mechanical supervision. Moreover, widely used datasets such as the Materials Project (MP) \citep{horton2025accelerated}, OQMD \citep{kirklin2015open}, and Alexandria \citep{schmidt2023machine} are derived largely from known or proposed compounds, introducing an implicit coverage bias toward already explored regions of crystal-structure space. Consequently, while diffusion pretraining reduces the dependence on quantum-mechanical labels, its scale and diversity remain constrained by the availability of suitable equilibrium structures. 

\begin{figure}[t]
    \centering
    \includegraphics[width=\linewidth]{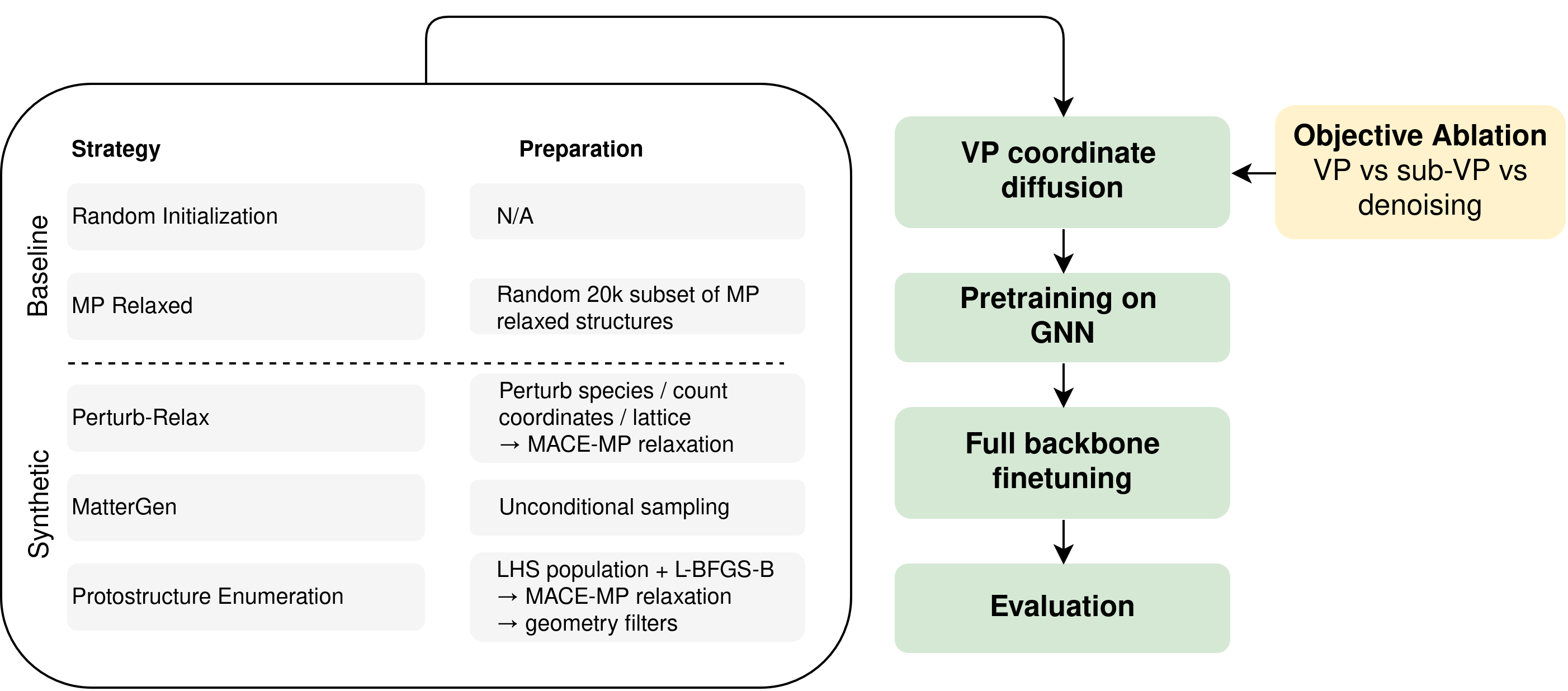}
    \caption{Overview of the dataset construction, pretraining, fine-tuning, and evaluation workflow. All data sources feed a common pretraining and evaluation protocol, with random initialization serving as the no-pretraining baseline.}
    \label{fig:workflow-overview}
\end{figure}

In this work, we address this bottleneck by coupling self-supervised pretraining with synthetic crystal structure generation. Our experimental design is summarized in Figure~\ref{fig:workflow-overview}. We first compare coordinate denoising \citep{zaidi_pre-training_2022} with continuous-time variance-preserving (VP) and sub-variance-preserving (sub-VP) diffusion objectives \citep{song_score-based_2020} across a set of MatBench property-prediction tasks \citep{dunn2020benchmarking}, selecting VP coordinate diffusion for the subsequent experiments. Using this objective, we compare pretraining on DFT-relaxed Materials Project structures with three complementary synthetic-data strategies: perturbing and relaxing generated structures, unconditional sampling from MatterGen \citep{zeni_generative_2025}, and protostructure enumeration \citep{parackal_screening_2026}. For strategies requiring structural relaxation, we replace per-structure DFT calculations with the MACE machine-learning interatomic potential (MLIP) \citep{batatia_mace_2022}, substantially reducing the cost of constructing and scaling the pretraining datasets. Following pretraining, the full GNN backbone is fine-tuned and evaluated on downstream property-prediction tasks. Across all seven source-comparison tasks, every synthetic-pretraining condition numerically improves over random initialization. The scaling study adds \texttt{mp\_gap} as an eighth task; under its fixed-epoch protocol, larger protostructure corpora approach or exceed MP-20k on most tasks while using both more structures and more optimization updates.

\section{Methods}

\paragraph{Self-supervised pretraining objectives}
We compare three self-supervised pretraining objectives: coordinate denoising \citep{zaidi_pre-training_2022}, variance-preserving (VP) diffusion, and sub-variance-preserving (sub-VP) diffusion \citep{song_score-based_2020}. For this comparison, all objectives are pretrained on the same random sample of 20,000 DFT-relaxed Materials Project (MP) structures, which we refer to throughout as \textbf{MP-20k}. Coordinate denoising perturbs equilibrium atomic coordinates with Gaussian noise and trains the model to recover the applied perturbation. For VP and sub-VP diffusion, structures are perturbed continuously at $t\sim\mathcal U(0,1)$ using a linear schedule with $\beta_{\min}=0.1$ and $\beta_{\max}=2$, and the model predicts mean-centered Cartesian noise with a mean-squared-error (MSE) loss. In all cases, only atomic coordinates are perturbed, while atomic species and lattice vectors remain fixed. We use this comparison to select the pretraining objective used in the subsequent synthetic-data experiments.

\paragraph{Pretraining data sources}
We compare MP-20k against three synthetic structure-generation strategies: 1) \textbf{Perturb-and-relax} randomly modifies the composition, coordinates (Gaussian standard deviation 1), and lattice of 30,000 MP parents, retaining 20,000 candidates after relaxation and filtering; 2) \textbf{MatterGen} samples \texttt{mattergen\_base} unconditionally with its default configuration and batch size 64, without subsequent filtering, deduplication, or relaxation \citep{zeni_generative_2025}; 3) \textbf{Protostructure enumeration} generates and populates symmetry-constrained labels following \citet{parackal_screening_2026} to construct a corpus of 165,000 viable synthetic structures. Allowed lattice and Wyckoff parameters are sampled by Latin hypercube sampling, and labels are populated with elements of atomic number below 90 excluding noble gases. Populated candidates are refined with L-BFGS-B, relaxed, filtered to 5--60~\AA$^3$/atom and minimum pair distance $\geq1.5$~\AA, then sorted by decreasing minimum distance to form exact nested datasets. Perturb-and-relax and protostructure candidates are optimized for at most 5,000 FIRE steps using \texttt{torch-sim-atomistic} 0.6.1 \citep{cohen2025torchsim} and the MACE-MP large model \citep{batatia_mace_2022}; coordinates and cells are optimized jointly, and failed or non-converged structures are discarded. No explicit fixed seeds were used for dataset sampling or MatterGen generation.

\paragraph{Model and pretraining setup}
We use Orb-v2 \citep{neumann_orb_2024} as the GNN backbone for all experiments. For all three self-supervised objectives, Orb-v2 is randomly initialized. VP and sub-VP additionally use a per-atom noise-prediction head and sinusoidal diffusion-time embeddings. Unless otherwise stated, each pretraining dataset contains 20,000 structures. The fixed-size data-source comparison therefore evaluates 20k structures from each synthetic source against MP-20k. To study whether synthetic generation can compensate for lower effectiveness per structure through scale, we additionally construct nested protostructure datasets containing 20k, 50k, 75k, 110k, and 165k structures.

\paragraph{Downstream evaluation}
After pretraining, the full Orb-v2 backbone is fine-tuned separately on MatBench property prediction tasks \citep{dunn2020benchmarking} using the same training procedure across all initialization conditions. The pretraining-objective comparison is evaluated on \texttt{matbench\_log\_gvrh}, \texttt{matbench\_jdft2d}, and \texttt{matbench\_dielectric}. The synthetic data-source comparison additionally includes \texttt{matbench\_log\_kvrh}, \texttt{matbench\_phonons}, \texttt{matbench\_2d}, and \texttt{matbench\_perovskites}, while the protostructure scaling study also evaluates \texttt{matbench\_mp\_gap}. We use only predefined MatBench fold 0; its training-and-validation portion is split 90/10 automatically, and the held-out test targets are never used for optimization or selection. Each fixed pretrained checkpoint is independently fine-tuned five times, and we report mean test MAE. Where run-level results were retained, standard error (SE) is the sample standard deviation of the five MAEs divided by $\sqrt{5}$; it captures fine-tuning variability from one checkpoint, not pretraining or dataset-selection variability. We compare against random initialization and MP-20k, with released Orb-v2 weights as an additional reference where relevant.

\paragraph{Optimization details}
Pretraining uses AdamW (learning rate $5\times10^{-5}$, zero weight decay), batch size 16, 400 epochs, float32 precision, and gradient-norm clipping at 1.0, with no scheduler, warmup, validation, or early stopping; the final checkpoint is used. Fine-tuning uses mean-squared-error loss and AdamW (learning rate $8\times10^{-5}$, weight decay 0.1, $\beta=(0.9,0.95)$, $\epsilon=10^{-8}$), batch size 6, and at most 500 epochs. A validation-MAE plateau scheduler uses factor 0.8, patience 3, and minimum learning rate $10^{-8}$; early stopping uses patience 50 and minimum improvement $10^{-8}$. Training uses float32 and elementwise gradient clipping at 1.0; inference uses bfloat16 and batch size 1. These settings follow the MatterTune Orb MatBench workflow and are fixed across tasks and initializations.

\paragraph{Computational cost}
We measured end-to-end dataset-construction time for 100 retained structures per source on a single NVIDIA A40 GPU, using four CPU threads and a batch size of 64. MatterGen required 13.20 minutes, protostructure generation required 39.36 minutes, and perturb-and-relax required 66.36 minutes. The respective pipelines attempted 100, 158, and 858 candidates. Timings include initialization, model loading, candidate generation, applicable screening and relaxation, rejected attempts, and output writing, but exclude environment setup and checkpoint downloads. MatterGen outputs were retained without additional relaxation or filtering. Both relaxation-based benchmarks used the same MACE checkpoint with TorchSim 0.4.0 (pinned for this timing benchmark; the pretraining datasets were relaxed with \texttt{torch-sim-atomistic} 0.6.1), float32 precision, joint atomic-position and cell relaxation, a force/cell-force tolerance of 0.03~eV/\AA, and a maximum of 5,000 FIRE steps.  Pretraining and fine-tuning each used a single NVIDIA A40 (48\,GB) GPU on a 64-CPU node. A 400-epoch VP pretraining run took 22.5\,h on MP-20k and 280.6\,h on the 165k protostructure corpus, and median fine-tuning time per run ranged from 0.2\,h (\texttt{jdft2d}) to 7.9\,h (\texttt{perovskites}), with \texttt{mp\_gap} requiring approximately 81\,h (Appendix Tables~\ref{tab:pretrain-time} and~\ref{tab:finetune-time}).

Our pipeline expands the synthetic pretraining corpus using existing pretrained models without performing new per-structure DFT calculations. Published studies provide context for the computational benefit of this substitution: \citet{kabakibo2026comparativestudymoleculardynamics} report an average 378-fold wall-clock acceleration for solid-electrolyte molecular dynamics using MACE on one A100 GPU relative to DFT on a 64-CPU node. Similarly, \citet{fu2025benchmarkquantumchemistryrelaxations} report that DFT geometry optimization of small organic molecules in PubChemQC can require several hours per molecule, with generation of approximately 3.5 million molecular relaxation trajectories taking several years across multiple high-performance computing systems. These literature results concern different systems, computational settings, and hardware; they support the motivation for avoiding additional DFT calculations but do not establish a quantitative speedup for our crystal-relaxation pipeline.

\paragraph{Code and existing assets} All custom code is publicly available: the Orb-v2 noise-head fork, the MatterTune diffusion-pretraining fork, and the protostructure generation pipeline; exact commits are listed in Appendix~\ref{app:software}. MP structures are distributed under CC BY 4.0; MatBench, MatterGen, MatterTune, MACE-MP, and TorchSim use the MIT License; and Orb uses Apache License 2.0. All are credited and used in accordance with their terms.
\section{Results}

\paragraph{Pretraining objective comparison}
We first compare coordinate denoising, VP diffusion, and sub-VP diffusion using the same MP-20k pretraining dataset.  Table~\ref{tab:sde-ablation} reports downstream MAE across three MatBench tasks. We additionally include two baselines: the released Orb-v2 model, which is fine-tuned from its original pretrained weights, and a randomly initialized Orb-v2 model, in which the pretrained weights are replaced with random initialization before fine-tuning. Both diffusion objectives numerically outperform coordinate denoising on all three tasks. Sub-VP achieves the lowest MAE on \texttt{jdft2d} and \texttt{log\_gvrh}, while VP performs best on \texttt{dielectric}. The differences between VP and sub-VP are small, however, and do not indicate a consistent advantage for either formulation across tasks. We therefore select VP for all subsequent experiments based on its greater observed training stability.

\begin{table}[h]
\centering
\normalsize
\caption{Downstream property-prediction MAE for different initialization and self-supervised pretraining strategies. Coordinate denoising, VP, and sub-VP are pretrained on MP-20k. Orb-v2 pretrained initializes from the released Orb-v2 weights, while Orb-v2 random initializes the same model architecture with random weights. Each entry reports the mean over five independent fine-tuning runs. Lower is better, and the best tested MP-20k pretraining objective in each column is bolded. Both diffusion objectives are numerically better than coordinate denoising on all three tasks.}
\label{tab:sde-ablation}
\sisetup{detect-weight=true, mode=text}
\renewcommand{\arraystretch}{1.2}

\begin{tabular}{
l
S[table-format=2.2]
S[table-format=1.3]
S[table-format=1.3]
}
\toprule
& \multicolumn{3}{c}{\textbf{MAE} ($\downarrow$)} \\
\cmidrule(lr){2-4}
\textbf{Initialization / Pretraining}
& {\texttt{jdft2d}}
& {\texttt{log\_gvrh}}
& {\texttt{dielectric}} \\
& {meV/atom}
& {$\log(\mathrm{GPa})$}
& {-} \\
\midrule
Coordinate denoising \citep{zaidi_pre-training_2022}
& 22.50 & 0.071 & 0.220 \\
VP
&  21.29 & 0.066 & \bfseries 0.190 \\
sub-VP
& \bfseries 20.60 & \bfseries 0.063 & 0.191 \\
\midrule
Orb-v2 pretrained
& 21.60 & 0.053 & 0.150 \\
Orb-v2 random
& 32.70 & 0.082 & 0.302 \\
\bottomrule
\end{tabular}
\end{table}

\paragraph{Synthetic pretraining data sources}
We next fix the pretraining dataset size at 20k structures and compare the three synthetic data sources against MP-20k and random initialization. Random-initialization results are available for all seven source-comparison tasks, and Table~\ref{tab:synth-strategies} shows that every synthetic-pretraining mean is numerically lower than its random-initialization counterpart. Averaging the task-wise relative MAE reductions computed from the displayed means across all seven tasks gives 31.8\% for MatterGen, 30.4\% for perturb-and-relax, and 30.9\% for protostructure enumeration, compared with 32.7\% for MP-20k.

At the same 20k dataset size, no single synthetic source performs best across all downstream tasks. MatterGen performs particularly well on \texttt{jdft2d} and closely matches MP-20k on the mechanical-property tasks, while perturb-and-relax achieves the best synthetic performance on \texttt{dielectric} and \texttt{perovskites}. Protostructure enumeration is less effective than the other synthetic sources on several tasks, most notably \texttt{jdft2d}, but performs strongly on \texttt{phonons} and \texttt{2d}. Overall, MatterGen and perturb-and-relax most closely approach the aggregate performance of MP-20k at a fixed pretraining size. Protostructure enumeration, however, offers a complementary advantage: structures can be generated systematically on demand, making dataset size readily scalable. This motivates the scaling study below.

\begin{table}[h]
\centering
\caption{Downstream property-prediction MAE across seven MatBench tasks for different 20k pretraining datasets. Results are mean $\pm$ SE over five fine-tuning runs where run-level values were retained; entries without $\pm$ report the available mean only. Lower is better, and the best synthetic mean in each task is bolded. Every synthetic-pretraining mean is numerically lower than random initialization, but no synthetic source is best on every task.}
\label{tab:synth-strategies}
\renewcommand{\arraystretch}{1.2}

\resizebox{\linewidth}{!}{%
\begin{tabular}{lccccccc}
\toprule
& \multicolumn{7}{c}{\textbf{Mean MAE $\pm$ SE} ($\downarrow$)} \\
\cmidrule(lr){2-8}
\textbf{Pretraining data}
& {\texttt{log\_gvrh}}
& {\texttt{log\_kvrh}}
& {\texttt{jdft2d}}
& {\texttt{dielectric}}
& {\texttt{perovskites}}
& {\texttt{phonons}}
& {\texttt{2d}} \\
\midrule

MatterGen
& $\mathbf{0.067 \pm {<}0.001}$
& $\mathbf{0.054 \pm {<}0.001}$
& $\mathbf{21.825 \pm 0.559}$
& $0.215 \pm 0.012$
& $0.032 \pm {<}0.001$
& $43.439 \pm 1.356$
& $\mathbf{0.120 \pm {<}0.001}$ \\

Perturb-and-relax
& $0.068 \pm {<}0.001$
& $0.065 \pm 0.003$
& $22.369 \pm 0.428$
& $\mathbf{0.189 \pm 0.009}$
& $\mathbf{0.031 \pm {<}0.001}$
& $44.546 \pm 2.041$
& $0.121 \pm 0.001$ \\

Protostructure
& $\mathbf{0.067 \pm {<}0.001}$
& $\mathbf{0.054 \pm {<}0.001}$
& $24.866 \pm 1.568$
& $0.213 \pm 0.013$
& $0.033 \pm {<}0.001$
& $\mathbf{39.618 \pm 2.205}$
& $0.122 \pm 0.001$ \\

\midrule

MP-20k
& $0.066 \pm {<}0.001$
& $0.054 \pm {<}0.001$
& $21.288 \pm 0.614$
& $0.190 \pm 0.009$
& $0.032 \pm {<}0.001$
& $43.670 \pm 1.933$
& $0.128 \pm 0.001$ \\

Random initialization
& $0.082$
& $0.076$
& $32.700$
& $0.302$
& $0.056$
& $65.430$
& $0.189$\\

\bottomrule
\end{tabular}%
}
\end{table}

\paragraph{Scaling protostructure pretraining data}
We therefore scale the protostructure dataset from 20k to 165k structures while keeping MP-20k as a fixed reference. Table~\ref{tab:protostructure-scaling} shows that downstream performance generally improves as the number of protostructure pretraining examples increases. The trend is particularly clear for \texttt{log\_gvrh}, \texttt{log\_kvrh}, \texttt{jdft2d}, \texttt{dielectric}, \texttt{perovskites}, and \texttt{2d}, although the improvement is not strictly monotonic for every task.

At 20k structures, protostructure pretraining underperforms MP-20k on most tasks. As the dataset grows, however, this gap narrows substantially. At 110k--165k structures, protostructure pretraining numerically matches or outperforms MP-20k on most of the evaluated tasks. For example, the 165k dataset improves MAE from 0.066 to 0.061 on \texttt{log\_gvrh}, from 0.054 to 0.049 on \texttt{log\_kvrh}, from 43.670 to 35.080 on \texttt{phonons}, and from 0.128 to 0.116 on \texttt{2d}. On \texttt{jdft2d} and \texttt{dielectric}, the best protostructure results occur at 110k structures and are comparable to the MP-20k reference. Because every corpus is trained for 400 epochs, increasing corpus size also increases the number of optimizer updates. These results therefore demonstrate joint data-and-compute scaling rather than isolating a pure data-volume effect.

The scaling trend is weaker for \texttt{mp\_gap}. MAE improves from 0.199 at 20k to 0.188 at 50k, reaches 0.187 at 110k, and then largely plateaus, suggesting that additional data and optimization updates do not provide the same benefit for all downstream tasks. This task contains substantially more labeled training examples than the other benchmarks considered here, consistent with the broader observation that the benefit of pretraining diminishes when downstream supervision is already abundant.

\begin{table}[h]
\centering
\caption{Downstream MAE as the protostructure pretraining set grows. Results are mean $\pm$ SE over five fine-tuning runs where retained; entries without $\pm$ report the available mean only. Lower is better, and the best protostructure mean in each column is bolded. MP-20k is the fixed DFT-relaxed reference. Larger corpora generally improve transfer, whereas \texttt{mp\_gap} improves initially and then plateaus.}
\label{tab:protostructure-scaling}
\renewcommand{\arraystretch}{1.2}

\resizebox{\linewidth}{!}{%
\begin{tabular}{lcccccccc}
\toprule
& \multicolumn{8}{c}{\textbf{Mean MAE $\pm$ SE} ($\downarrow$)} \\
\cmidrule(lr){2-9}
\textbf{Size}
& {\texttt{log\_gvrh}}
& {\texttt{log\_kvrh}}
& {\texttt{jdft2d}}
& {\texttt{dielectric}}
& {\texttt{perovskites}}
& {\texttt{phonons}}
& {\texttt{2d}}
& {\texttt{mp\_gap}} \\
\midrule

20k
& $0.067 \pm {<}0.001$
& $0.054 \pm {<}0.001$
& $24.866 \pm 1.568$
& $0.213 \pm 0.013$
& $0.033 \pm {<}0.001$
& $39.618 \pm 2.205$
& $0.122 \pm 0.001$
& $0.199 \pm 0.001$ \\

50k
& $0.064 \pm {<}0.001$
& $0.051 \pm {<}0.001$
& $23.186 \pm 0.815$
& $0.208 \pm 0.012$
& $0.030 \pm {<}0.001$
& $39.188 \pm 2.191$
& $0.120 \pm {<}0.001$
& $0.188 \pm {<}0.001$ \\

75k
& $0.062 \pm {<}0.001$
& $0.051 \pm {<}0.001$
& $21.372 \pm 0.872$
& $0.198 \pm 0.013$
& $0.030 \pm {<}0.001$
& $40.710 \pm 1.164$
& $\mathbf{0.116 \pm {<}0.001}$
& $0.191 \pm 0.002$ \\

110k
& $0.062 \pm {<}0.001$
& $0.050 \pm {<}0.001$
& $\mathbf{20.330 \pm 0.596}$
& $\mathbf{0.183 \pm 0.008}$
& $\mathbf{0.029 \pm {<}0.001}$
& $37.756 \pm 0.479$
& $0.119 \pm {<}0.001$
& $\mathbf{0.187 \pm {<}0.001}$ \\

165k
& $\mathbf{0.061 \pm {<}0.001}$
& $\mathbf{0.049 \pm {<}0.001}$
& $21.716 \pm 0.893$
& $0.188 \pm 0.010$
& $\mathbf{0.029 \pm {<}0.001}$
& $\mathbf{35.080 \pm 1.469}$
& $\mathbf{0.116 \pm {<}0.001}$
& $0.188 \pm 0.001$\\

\midrule

MP-20k
& $0.066 \pm {<}0.001$
& $0.054 \pm {<}0.001$
& $21.288 \pm 0.614$
& $0.190 \pm 0.009$
& $0.032 \pm {<}0.001$
& $43.670 \pm 1.933$
& $0.128 \pm 0.001$
& $0.196 \pm 0.002$\\

\bottomrule
\end{tabular}%
}
\end{table}

\begin{figure}[h]
\centering
\includegraphics[width=.85\linewidth]{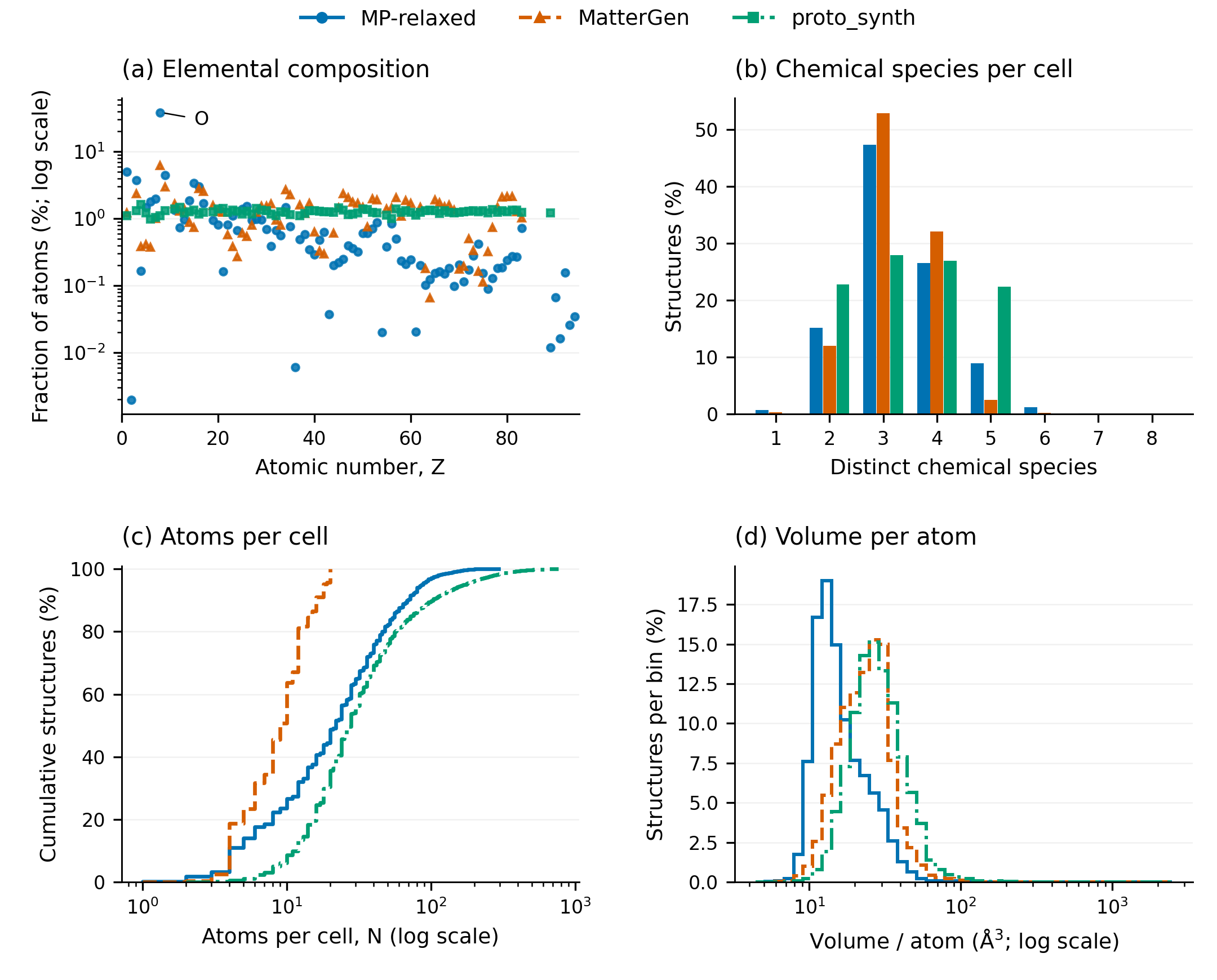}
\caption{Structural and compositional distributions of the three pretraining sources, each containing 20,000 structures.
(a) Fraction of all atoms belonging to each element, shown on a logarithmic vertical axis; absent elements are omitted.
(b) Percentage of structures containing each number of distinct chemical species.
(c) Empirical cumulative distributions of atom counts in the supplied cells, with a logarithmic horizontal axis.
(d) Volume-per-atom distributions using common logarithmically spaced bins, expressed as percentages of structures per bin.}

\label{fig:comparison}
\end{figure}

\paragraph{Pretraining data distributions}
To better understand the differences among pretraining sources, Figure~\ref{fig:comparison} compares the elemental composition, number of distinct species, atom count, and volume per atom of the three 20k data sets. MP-20k is strongly oxygen-dominated, with oxygen accounting for 36.6\% of atoms, compared to 6.2\% for MatterGen and 1.1\% for the protostructure dataset. The protostructure dataset exhibits a more uniform elemental distribution across its represented elements, rather than a large number of distinct elements. It also samples two to five species structures more evenly, whereas three-species structures dominate MP-20k and MatterGen.

The three datasets also differ geometrically. MatterGen contains at most 20 atoms per cell, with a median of 9, compared with medians of 22 and 28 for MP-20k and protostructure dataset respectively. Median volumes per atom are 14.6, 23.8 and 27.4 ~\AA$^{3}$ respectively, indicating differences beyond cell size alone. These observations suggest that protostructure pretraining allocates its fixed sample budget differently across chemical and structural space. Less concentrated sampling possibly contributes to the weaker performance of the protostructure dataset at 20k examples, while larger datasets may improve coverage of that space. This interpretation is consistent with the scaling results; however, this does not establish a direct explanation.

\FloatBarrier
\section{Discussion and conclusion}

Our results show that synthetic crystal structures can provide numerically effective pretraining data for self-supervised materials GNNs without requiring new DFT relaxations for every structure. At a fixed dataset size of 20k, MatterGen and perturb-and-relax most closely approach the performance of MP-20k, while protostructure enumeration is generally less effective per structure. Protostructure enumeration offers a complementary advantage because its corpus can be expanded systematically without being limited by the size of an existing DFT-relaxed database. Under our fixed 400-epoch protocol, scaling this corpus substantially narrows the performance gap, with larger synthetic datasets numerically matching or exceeding MP-20k on most evaluated tasks. This is a data-plus-compute result: larger corpora provide both more distinct structures and proportionally more optimization updates.

These results highlight a tradeoff between per-structure effectiveness and scalability. Synthetic sources whose distributions more closely resemble MP data can provide strong transfer performance at relatively small dataset sizes, whereas broader enumerated datasets may require more examples and/or optimization to achieve comparable performance. The latter can be generated at scale and relaxed using MLIPs rather than DFT, making substantially larger pretraining corpora practical. The current experiment does not separate the effect of additional structural diversity from that of additional optimizer updates.

Overall, our findings support synthetic structure generation as a viable route toward scalable self-supervised pretraining for materials property prediction. Every synthetic condition numerically outperforms random initialization across the seven source-comparison tasks, and the eight-task scaling study shows competitive transfer from a large protostructure corpus. Compute-matched experiments are needed before attributing the scaling gains to data volume or diversity alone.

\paragraph{Limitations}
Our study has several limitations. First, the experiments use a single GNN backbone, one MatBench fold, and one selected pretrained checkpoint per condition, so the observed trends may not generalize uniformly across folds, pretraining seeds, checkpoint choices, model architectures, or downstream properties. Second, protostructure enumeration exhibits lower effectiveness per structure than MP-20k and the other synthetic sources at the fixed 20k dataset size. Because all corpora are trained for 400 epochs, the 165k run processes 8.25 times as many structure presentations as the 20k run; dataset size is therefore confounded with optimization compute. Improving how compositions, prototypes, and structural parameters are sampled may reduce the efficiency gap, but compute-matched scaling is needed to isolate the contribution of additional data.

The benefit of scaling also varies across downstream tasks. Improvements are clearest for several of the smaller MatBench datasets, whereas the trend is weak for the comparatively data-rich \texttt{mp\_gap} task. This suggests that additional pretraining data are most beneficial when downstream supervision is limited, while their marginal value may decrease as the size of the labeled fine-tuning dataset grows. Finally, although MLIP relaxation avoids new DFT calculations, generating and relaxing large synthetic datasets still incurs non-negligible computational cost and may inherit inaccuracies or biases from the MLIP used for relaxation.

\paragraph{Broader impacts}
More scalable pretraining data can help reduce the reliance on expensive DFT calculations, broadening access to materials ML and accelerating candidate screening for energy, catalysis and electronic applications. On the other hand, biases or errors from source databases and generators can produce misleading predictions and waste computational resources. Synthetic candidates should therefore be validated with higher-fidelity calculations and must be verified before consequential use.

\paragraph{Future work}
Future work should investigate how the composition and structural diversity of synthetic pretraining datasets affect transfer performance more directly. In particular, more targeted protostructure sampling could balance broad coverage of crystal-structure space with sufficient representation of frequently occurring chemical environments. Compute-matched scaling studies should separate generation strategy and dataset size from the number of optimization updates, including for generative sources such as MatterGen. Finally, evaluating synthetic-data pretraining across additional model architectures, downstream tasks, folds, and pretrained seeds would help determine how broadly these trends generalize.

\bibliographystyle{plainnat}
\bibliography{references}

\begin{theappendices}

\chapter{Software and repository versions}
\label{app:software}
The custom code used for diffusion pretraining, synthetic structure generation, and downstream evaluation is publicly available at the repositories and commits listed in Table~\ref{tab:software}.

\begin{table}[H]
  \centering
  \caption{Software provenance for the custom and generative components used in this study. Commits are abbreviated to 12 characters.}
  \label{tab:software}
  \small
  \begin{tabularx}{\linewidth}{@{}p{2.3cm}Xp{2.3cm}p{3.0cm}@{}}
    \toprule
    Component & Repository & Commit & Role \\
    \midrule
    Orb noise-head fork & \url{https://github.com/aamodvarma/orb-models-noisehead} & \texttt{8d5b40059e48} & Diffusion backbone and noise head \\
    MatterTune fork & \url{https://github.com/aamodvarma/MatterTune_diffusion} & \texttt{f2a464b3e97b} & Pretraining and downstream fine-tuning \\
    Protostructure code & \url{https://github.com/aamodvarma/protostructure} & \texttt{9464e1457135} & Enumeration, population, and screening \\
    MatterGen & \url{https://github.com/microsoft/mattergen} & \texttt{ac9ddd406171} & Unconditional structure generation \\
    \bottomrule
  \end{tabularx}
\end{table}

MatterTune and its dependencies were installed following the MatterTune installation guide. A complete environment lock file was not retained for these experiments.

\clearpage
\section{Python implementation details}
\label{app:python_details}

The predefined MatBench training-and-validation portion was obtained with \path{get_train_and_val_data}, and the held-out test portion was obtained with \path{get_test_data}.

The per-atom noise-prediction head was implemented with the \verb|NoiseHead| class.

MatterGen sampling was invoked from the command line using the default sampling configuration.

Protostructure labels were repeatedly enumerated and populated until the relaxed and filtered pool contained at least 165,000 viable structures. Sampling used \verb|LatinHypercube| from \verb|SciPy|. A stratified permutation sampler provided the implementation fallback. The decreasing-minimum-distance ordering was selected with \verb|--sort-by min_dist|.

The MACE-MP large model was loaded with \path{mace_mp(model="large", return_raw_model=True)} and wrapped as a TorchSim \verb|MaceModel|.

The plateau-based learning-rate scheduler was implemented with \verb|ReduceLROnPlateau|.

\section{Training timings}

\begin{table}[h]
  \centering
  \caption{Wall-clock VP-diffusion pretraining time for the final checkpoint of each condition (400 epochs, batch size 16). All runs used a single NVIDIA A40 (48\,GB) GPU on a 64-CPU node; each condition was pretrained once.}
  \label{tab:pretrain-time}
  \begin{tabular}{lcc}
    \toprule
    Pretraining data & Wall-clock (h) & Min / epoch \\
    \midrule
    MP-20k                  & 22.5  & 3.4  \\
    Perturb-and-relax (20k) & 27.2  & 4.1  \\
    MatterGen (20k)         & 29.8  & 4.5  \\
    Protostructure 20k      & 48.6  & 7.3  \\
    Protostructure 50k      & 134.1 & 20.1 \\
    Protostructure 75k      & 134.3 & 20.2 \\
    Protostructure 110k     & 183.9 & 27.6 \\
    Protostructure 165k     & 280.6 & 42.1 \\
    \bottomrule
  \end{tabular}
\end{table}
\begin{table}[h]
  \centering
  \caption{Wall-clock fine-tuning time per run on MatBench fold 0. All runs used a single NVIDIA A40 (48\,GB) GPU on a 64-CPU node; only runs that completed normally are included. Mean epochs is the number of epochs completed before early stopping or the epoch limit.}
  \label{tab:finetune-time}
  \begin{tabular}{lccc}
    \toprule
    Task & Mean (h) & Median (h) & Mean epochs \\
    \midrule
    \texttt{jdft2d}      & 0.27  & 0.18  & 92  \\
    \texttt{phonons}     & 0.59  & 0.52  & 158 \\
    \texttt{dielectric}  & 1.89  & 1.40  & 100 \\
    \texttt{2d}          & 2.05  & 2.03  & 147 \\
    \texttt{log\_kvrh}   & 2.39  & 2.32  & 80  \\
    \texttt{log\_gvrh}   & 4.32  & 3.38  & 121 \\
    \texttt{perovskites} & 7.97  & 7.91  & 166 \\
    \texttt{mp\_gap}     & 82.33 & 80.66 & 228 \\
    \bottomrule
  \end{tabular}
\end{table}
\end{theappendices}

\end{document}